\documentclass[letterpaper, 10pt, conference]{ieeeconf}

\IEEEoverridecommandlockouts
\usepackage{graphicx}
\usepackage{amsmath}
\usepackage{amssymb}
\usepackage{mathnotation}
\usepackage{cite}
\usepackage{mathtools}
\usepackage{booktabs}
\newcommand{\censor}[1]{#1}

\title{\LARGE \bf
Coordinate-Independent Robot Model Identification
}

\author{Yanhao Yang, Ross L. Hatton %
\thanks{This work was supported in part by NSF Grants No. 1653220, 1826446, and 1935324, and by ONR Grant No. N00014-23-1-2171.} %
\thanks{Y. Yang and R. L. Hatton are with the  Collaborative Robotics and Intelligent Systems (CoRIS) Institute at Oregon State University, Corvallis, OR USA. {\tt\small \{yangyanh, Ross.Hatton\}@oregonstate.edu}} %
}

\begin{document}

\maketitle
\thispagestyle{empty}
\pagestyle{empty}

\begin{abstract}
Robot model identification is commonly performed by least-squares regression on inverse dynamics, but existing formulations measure residuals directly in coordinate force space and therefore depend on the chosen coordinate chart, units, and scaling. This paper proposes a coordinate-independent identification method that weights inverse-dynamics residuals by the dual metric induced by the system Riemannian metric. Using the force--velocity vector--covector duality, the dual metric provides a physically meaningful normalization of generalized forces, pulling coordinate residuals back into the ambient mechanical space and eliminating coordinate-induced bias. The resulting objective remains convex through an affine-metric and Schur-complement reformulation, and is compatible with physical-consistency constraints and geometric regularization. Experiments on an inertia-dominated Crazyflie--pendulum system and a drag-dominated \censor{Land}Salp robot show improved identification accuracy, especially on shape coordinates, on both downsampled and full datasets.
\end{abstract}

\section{Introduction}

Accurate robot dynamic models are central to modern robotics. They underpin model-based control \cite{ref171}, physical simulation for robot learning \cite{ref525}, state estimation \cite{ref029}, trajectory optimization \cite{ref557}, and hardware design \cite{ref451}, especially in regimes where performance, safety, and sample efficiency matter. Whether the goal is dexterous manipulation, agile locomotion, marine propulsion, or interaction with uncertain environments, model quality strongly affects downstream performance \cite{ref270}. In practice, however, first-principles models are rarely exact: unmodeled friction, fluid effects, compliance, manufacturing tolerances, and parameter uncertainty all create a gap between nominal and real behavior \cite{ref014}. Robot model identification therefore remains a fundamental problem, providing a systematic way to estimate dynamic parameters from data and reduce this modeling gap \cite{ref327}.

A broad range of identification methods has been developed in robotics. Classical approaches include frequency-response identification \cite{ref429}, least-squares fitting of inverse dynamics \cite{ref024}, and residual fitting on nominal models \cite{ref596}. More recent data-driven methods incorporate geometric mechanics \cite{ref041}, dynamic mode decomposition \cite{ref070}, self-supervised learning \cite{ref355}, neural differential models \cite{ref096}, and physics-informed learning \cite{ref198}. 

Among these methods, inverse-dynamics least squares remains one of the most widely used tools in practice because it leverages physical priors, often reduces identification to a linear or convex regression problem, scales well to large datasets, avoids repeated forward simulation, and integrates naturally with constraints and priors \cite{ref327}. It is a general approach for estimating rigid-body inertial parameters, and analogous formulations also arise in systems dominated by other generalized linear constitutive effects \cite{ref583,ref586}.

\begin{figure}[!t]
\centering
\includegraphics[width=\linewidth]{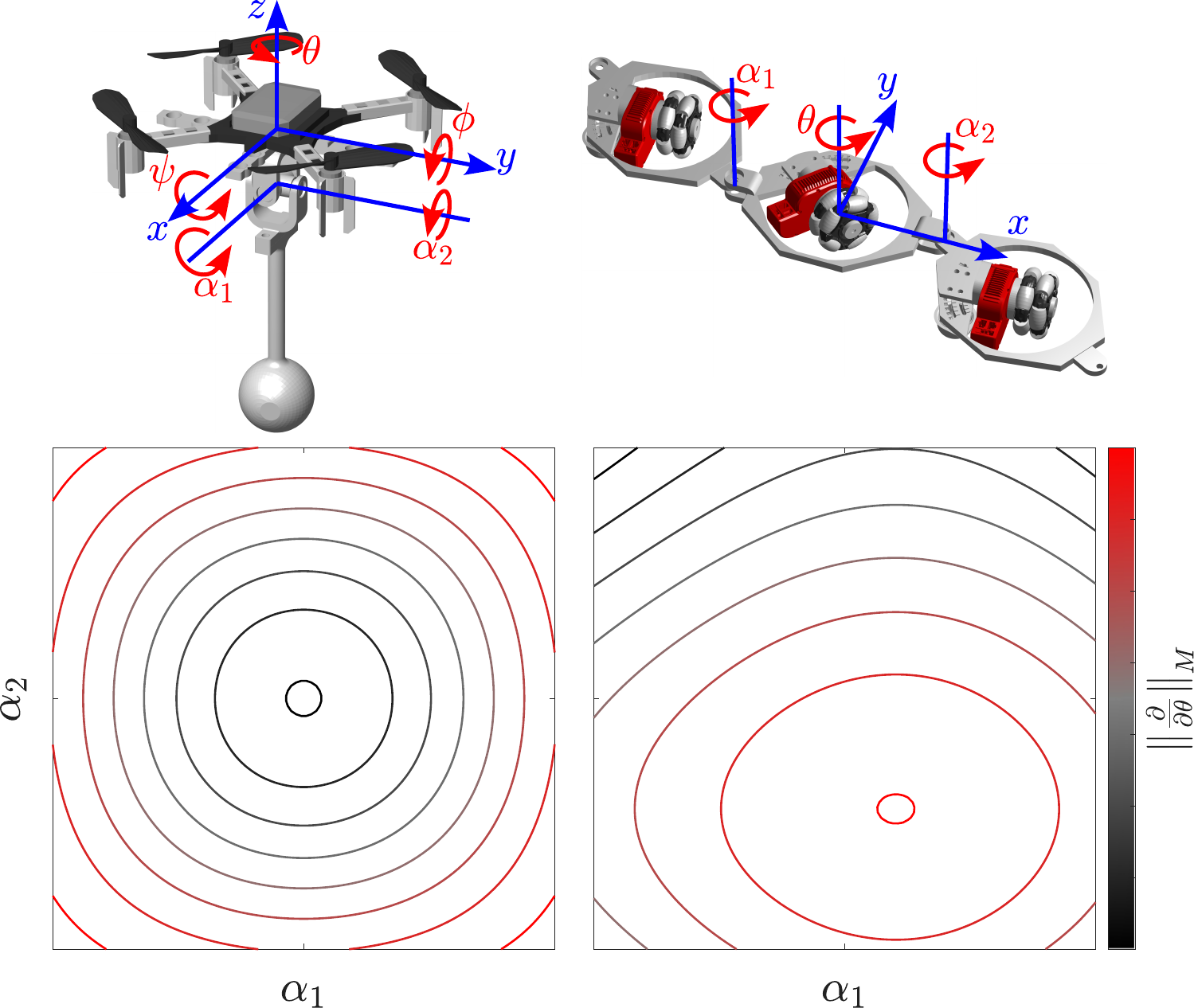}
\caption{Two robotic systems studied in this paper for model identification: an inertia-dominated system (left) and a drag-dominated system (right). In both systems, the generalized coordinates have different units and scales, and the magnitude of the inner product of a unit coordinate velocity under the corresponding Riemannian metric depends on the configuration, as illustrated in the bottom-left and bottom-right panels, which plot the inner-product value along the \( \dot{\theta} \) direction for the two systems, respectively. Together, these factors illustrate the coordinate dependence in the system identification problem addressed in this paper.}
\label{fig:merge}
\end{figure}

The key limitation of inverse-dynamics least squares, however, is geometric rather than computational. Its standard formulation can be interpreted as a maximum-likelihood estimator under independent, isotropic Gaussian residual assumptions, which leads to minimizing a Euclidean norm of residuals in coordinate generalized force space. This assumption implicitly treats all coordinates as commensurate and measures error according to the chosen coordinate chart rather than the underlying mechanics. Consequently, the identified model can depend on the units, scaling, or parameterization of the generalized coordinates. Coordinates expressed in radians, meters, or rescaled internal variables may contribute very differently to the objective even when they represent physically comparable effects. This issue is especially pronounced in systems with heterogeneous coordinates, multiple scales, or configuration-dependent metric distortion, such as the two systems in Fig.~\ref{fig:merge}. In such cases, Euclidean residual minimization can bias the fit toward coordinates with numerically larger force magnitudes, leading to models that fit one coordinate representation well but generalize poorly across equivalent reparameterizations.

Several extensions improve inverse-dynamics least squares, but they do not fully resolve coordinate dependence. Weighted least squares based on error covariance accounts for differences in coordinate units and scales; regularization improves conditioning and reduces overfitting in data-limited settings \cite{ref325}; physically consistent constraints enforce realizable inertial parameters \cite{ref554}; and energy-based residuals attempt to better reflect physical structure \cite{ref185}. These methods improve robustness, stability, and physical plausibility, but their weightings are typically introduced for statistical, numerical, or heuristic reasons rather than from the intrinsic dual structure of mechanics. As a result, the residual objective remains coordinate dependent. Energy-based objectives partially avoid direct force-residual minimization, but they also compress the information in inverse-dynamics equations and can reduce identifiability. Thus, a principled treatment of coordinate independence in inverse-dynamics identification is still lacking.

This paper addresses that gap by introducing a dual-metric-weighted least-squares formulation for inverse dynamics. For robot systems whose dynamics are governed by a Riemannian metric on the configuration manifold, generalized velocities are tangent vectors and generalized forces are cotangent vectors (``covectors'') in the cotangent space dual to the tangent space. The natural metric on force space is therefore not the implicit Euclidean metric used in standard least squares, but the dual metric induced by the Riemannian metric on velocities. Weighting inverse-dynamics residuals by this dual metric yields a physically meaningful normalization rooted in the force--velocity vector--covector duality through power. In effect, this pulls coordinate force residuals back into the ambient mechanical space, eliminating dependence on arbitrary coordinate scaling and reparameterization while preserving the computational advantages of inverse-dynamics least squares. The resulting problem can still be formulated as a convex optimization problem and remains compatible with standard enhancements such as physically consistent constraints and geometric regularization.

We validate the proposed method on both inertia-dominated and drag-dominated robotic systems (Fig.~\ref{fig:merge}). Across both platforms, the proposed dual-metric weighting improves model accuracy, generalization, and data efficiency relative to error-covariance-weighted least squares, energy-based objectives, and regularized least-squares methods with additional prior information. The advantage is most pronounced on shape coordinates, where the nonlinear coordinate representation of the configuration manifold makes coordinate dependence especially harmful, and it persists across both downsampled and full datasets.

\section{Preliminaries}

This section summarizes the standard inverse-dynamics least-squares formulation for robot system identification and the main extensions relevant to this paper.

\subsection{Inverse-Dynamics Least-Squares System Identification}

We consider robot systems whose dynamics can be written in inverse-dynamics form, i.e., as a map from state, state derivatives, and model parameters to applied generalized effort. Let \( \bundlespace \) denote the configuration manifold, \( \bundle \in \bundlespace \) the configuration, \( \bundledot \in T_\bundle \bundlespace \) generalized velocities, \( \bundleddot \) generalized accelerations, \( \lagrangeforce \in T_\bundle^*\bundlespace \) generalized forces (or efforts), and \( \dynpar \in \euclid^\dynpardim \) the dynamic parameter vector. The key assumption is that the inverse dynamics is affine in the unknown parameter vector \( \dynpar \), which enables least-squares identification.

In a chosen coordinate chart and its associated dual co-basis,\footnote{The coordinate coefficients of $\bundledot$ and $\bundleddot$ are represented as column vectors in $\mathbb{R}^n$, whereas the coordinate coefficients of $\lagrangeforce$ form a row vector under the natural pairing $\lagrangeforce(\bundledot)$.} this affine dependence yields, for sample \( \sampleidx \),
\begin{equation}
\lagrangeforce_\sampleidx
=
\regressor(\bundle_\sampleidx,\bundledot_\sampleidx,\bundleddot_\sampleidx)\dynpar + \varepsilon_\sampleidx,
\label{eq:ls}
\end{equation}
where \( \regressor(\cdot) \in \euclid^{\gendim \times \dynpardim} \) is the inverse-dynamics regressor, \( \gendim \) is the number of generalized velocities, and \( \varepsilon_\sampleidx \) collects measurement noise, unmodeled effects, and model mismatch. The ordinary least-squares (OLS) estimate is then obtained by minimizing the squared inverse-dynamics residual:
\begin{equation}
\hat{\dynpar}
=
\arg\min_{\dynpar}
\;
\sum_{\sampleidx=1}^{\samplenum}
\left\|
\regressor_\sampleidx \dynpar - \lagrangeforce_\sampleidx
\right\|_2^2.
\end{equation}
This formulation is attractive because it avoids repeated forward simulation, preserves sample-wise separability, and reduces system identification to a convex regression problem whenever the inverse dynamics is linear in the parameters.

This affine-in-parameters structure arises in multiple physical regimes. For rigid-body systems dominated by inertia, such as the Crazyflie--pendulum system in the top left of Fig.~\ref{fig:merge}, the inverse dynamics commonly takes the form
\begin{equation}\label{eq:inertia}
\lagrangeforce = M(\bundle)\bundleddot + C(\bundle,\bundledot)\bundledot + g(\bundle),
\end{equation}
where \( M(\bundle) \) is the mass matrix, \( C(\bundle,\bundledot)\bundledot \) collects Coriolis and centrifugal effects, and \( g(\bundle) \) is the generalized gravity term. A standard result in robot dynamics is that this model can be written linearly in a set of inertial parameters, where \( \dynpar \) may include link masses, center-of-mass moments, and inertia tensor terms.

In other robotic systems, inertial effects may be negligible relative to drag. In low-Reynolds-number locomotion, highly damped mechanisms, or systems moving through a resistive medium, such as the \censor{Land}Salp robot in the top right of Fig.~\ref{fig:merge}, a reduced-order model may be written as
\begin{equation}\label{eq:drag}
\lagrangeforce = M(\bundle)\bundledot,
\end{equation}
where \( M(\bundle) \) is a drag or resistance matrix. If the drag model is linear in the unknown drag coefficients, then it also admits the affine form in \eqref{eq:ls}, with \( \dynpar \) encoding those coefficients.

Across inertia-dominated, drag-dominated, and mixed regimes, the key requirement is not a particular physical mechanism, but rather that the inverse dynamics can be expressed as a regressor linear in the unknown parameters.

\subsection{Improvements to Inverse-Dynamics Least Squares}

Although OLS is widely used, several extensions have been proposed to improve robustness, physical plausibility, or statistical efficiency.

\subsubsection{Weighted Least Squares Using Error Covariance}

If the residual \( \noise \) is heteroscedastic or correlated, OLS is no longer statistically efficient. A standard generalization is weighted least squares (WLS),
\begin{equation}
\hat{\dynpar}
=
\arg\min_{\dynpar}
\;
\sum_{\sampleidx=1}^{\samplenum}
\transpose{\left(\regressor_\sampleidx\dynpar - \lagrangeforce_\sampleidx\right)}
\lsweight
\left(\regressor_\sampleidx\dynpar - \lagrangeforce_\sampleidx\right),
\end{equation}
where typically \( \lsweight = \inv{\noisecov} \) with residual covariance
\begin{equation}
\noisecov
=
\sum_{\sampleidx=1}^{\samplenum}
\mathbb{E}\!\left[\noise_\sampleidx \transpose{\noise}_\sampleidx\right]
\succ 0.
\end{equation}
When the true conditional residual covariance is known and the standard linear-regression assumptions hold, WLS is statistically optimal in the usual mean-squared-error sense. In inverse-dynamics identification, however, residuals also reflect noise propagated through nonlinear state reconstruction and differentiation, as well as model mismatch. Their effective covariance may therefore be state dependent and difficult to estimate reliably. Consequently, a fixed covariance weight defined in a particular coordinate chart need not eliminate the coordinate dependence of the residual.

\subsubsection{Physically Consistent Constraints}

A least-squares fit may yield parameters that explain measured data but violate physical realizability. For rigid-body identification, examples include nonpositive mass, non-positive-definite inertia tensors, or inertial parameters inconsistent with any realizable mass distribution \cite{ref554}. To address this problem, identification can be formulated as a constrained optimization problem:
\begin{equation}
\hat{\dynpar}
=
\arg\min_{\dynpar}
\;
\sum_{\sampleidx=1}^{\samplenum}
\|\regressor_\sampleidx\dynpar - \lagrangeforce_\sampleidx\|_2^2
\quad
\text{s.t.}
\quad
P(\dynpar) \succ 0,
\end{equation}
where, for mass--inertia parameters, \( P(\dynpar) \) is the pseudo-inertia matrix, and the constraint enforces positive mass and density-realizable moments of inertia; for viscous drag coefficients, \( P(\dynpar) \) is the drag coefficient matrix, and the constraint enforces nonnegative drag-induced dissipation.

When these constraints are expressed as linear matrix inequalities, the resulting problem is a convex semidefinite program. Such physically consistent identification improves interpretability, extrapolation, and compatibility with downstream control and simulation.

\subsubsection{Geometric Convex Regularization}

When data are limited or the regressor is ill-conditioned, the least-squares estimate may have high variance or be vulnerable to noise or outliers. Regularization addresses this problem by penalizing undesirable parameter values or deviations from prior structure \cite{ref325}. A general regularized problem is
\begin{equation}
\hat{\dynpar}
=
\arg\min_{\dynpar}
\;
\sum_{\sampleidx=1}^{\samplenum}
\|\regressor_\sampleidx\dynpar - \lagrangeforce_\sampleidx\|_2^2
+
\regscale \reg(\dynpar),
\end{equation}
where \( \reg(\dynpar) \) is a regularizer and \( \regscale \ge 0 \) is a tuning parameter.

Convex approximations to the geodesic distance between two positive-definite matrices can be used as regularizers. Two common examples are the Bregman divergence and the constant pullback metric evaluated at a nominal value \( \pi_0 \), obtained from CAD or a previously identified model. Such geometric convex regularization can improve conditioning, reduce overfitting, and encode structured priors while preserving convexity. In practice, however, these methods depend on the availability and quality of prior information: the nominal value \( \pi_0 \) may be unavailable or inaccurate due to modeling simplifications, manufacturing tolerances, or wear. Moreover, the regularization strength \( \regscale \) must be tuned, and performance may be sensitive to that choice.

\subsubsection{Energy-Related Residuals}

Another line of work modifies the residual so that the objective reflects the system's energetic structure \cite{ref185}. This approach uses an integrated power model relating work over a time interval to the corresponding energy change, avoiding explicit dependence on measured accelerations. Inspired by this idea, one can define a coordinate-invariant instantaneous power residual by pairing the force residual with the measured velocity:
\begin{equation}
\hat{\dynpar}
=
\arg\min_{\dynpar}
\;
\sum_{\sampleidx=1}^{\samplenum}
\left(
\transpose{\bundledot}_\sampleidx
\left(
\regressor_\sampleidx\dynpar - \lagrangeforce_\sampleidx
\right)
\right)^2.
\end{equation}
This formulation replaces the vector-valued force residual with a scalar power residual, thereby relating the identification objective to a physically meaningful quantity. For drag-dominated systems, it measures the mismatch between predicted drag-dissipation and measured actuator power.

However, because this residual compresses the vector-valued inverse-dynamics error into a scalar quantity, it may discard information contained in the full force balance and thereby reduce parameter identifiability.

\section{Methodology}

\begin{figure*}[!t]
\centering
\includegraphics[width=\linewidth]{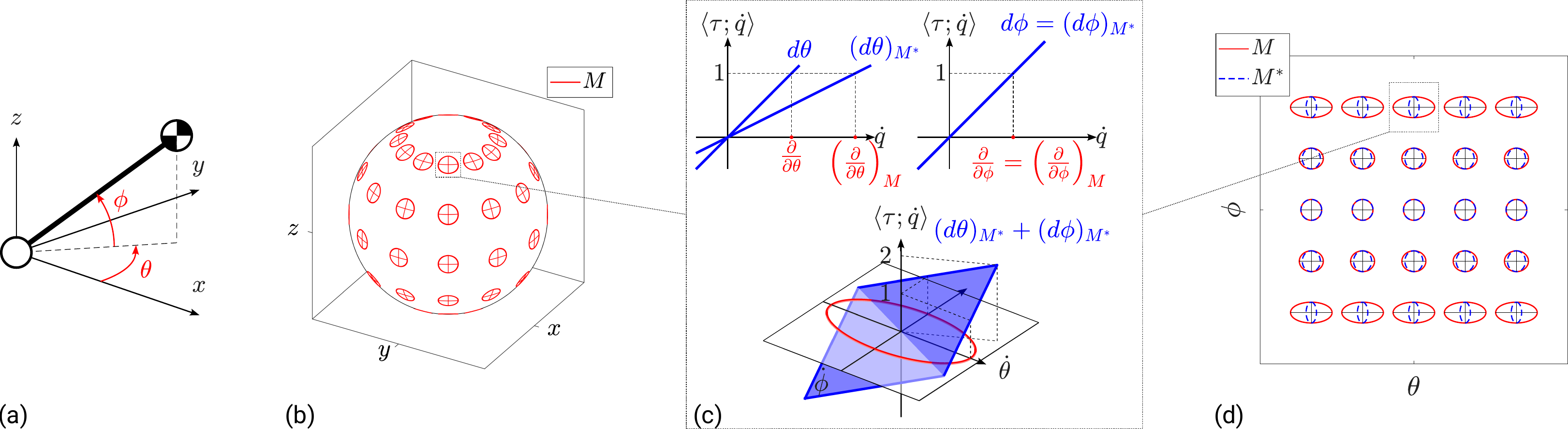}
\caption{Metric and dual-metric normalization illustrated by a pan--tilt mechanism. (a) A pan--tilt mechanism with a massless link and a point mass at the end effector; its generalized coordinates parameterize the configuration manifold through a nonlinear chart. (b) The Riemannian metric induced by the point-mass motion constrained by the mechanism and measured in the ambient Euclidean space. Red ellipses denote unit-\(\metric\)-norm velocity sets; their intercepts indicate metric-normalized tangent directions at each configuration. (c) Generalized velocities are tangent vectors, and generalized forces are covectors; their pairing produces power. A force covector specifies the rate of power production associated with motion in a given velocity direction. Through this vector--covector duality, the velocity metric canonically induces the corresponding dual metric on generalized-force space. (d) Coordinate representations of the velocity metric and its induced dual metric, visualized by red and blue ellipses representing unit velocity and unit force norms, respectively.}
\label{fig:idea}
\end{figure*}

We now develop a coordinate-independent inverse-dynamics identification objective using the duality between generalized velocities and generalized forces. Generalized coordinates live in a chart of the configuration manifold, while the physical motion and effort they represent live in an underlying ambient mechanical space. The system Riemannian metric provides the natural way to measure velocities in that space, and the induced dual metric provides the corresponding physically meaningful way to measure force residuals. We then show that, although the resulting weighted objective depends on the unknown dynamic parameters, it can still be reformulated as a convex optimization problem.

\subsection{Dual-Metric Weighting for Coordinate-Independence}

Suppose the system is equipped with a symmetric positive-definite metric tensor \( \metric(\bundle,\dynpar) \in \mathbb{S}_{++}^{\gendim} \), which defines the inner product and associated norm on the tangent space,
\begin{equation}
\langle \bundledot_1, \bundledot_2 \rangle_{\metric}
:=
\transpose{\bundledot_1}\metric\bundledot_2, \qquad \|\bundledot\|_{\metric}
:=
\sqrt{\langle \bundledot, \bundledot \rangle_{\metric}}.
\end{equation}
Geometrically, \( \metric \) measures generalized velocity after mapping coordinates into ambient mechanical space, making coordinate velocities physically commensurate. This interpretation is shown by the pan--tilt system in Fig.~\ref{fig:idea}(a). Let
\begin{equation}
\bundle = \begin{bmatrix}\theta \\ \phi\end{bmatrix},
\end{equation}
where \( \theta \) and \( \phi \) denote pan and tilt. Let \( x(\bundle) \in \euclid^3 \) denote the end-effector position. Its ambient velocity is
\begin{equation}
\dot{x} = J(\bundle)\bundledot,
\end{equation}
where \( J(\bundle) \) is the Jacobian. For a unit point mass at the end effector in Euclidean ambient space, the induced metric on generalized velocities is
\begin{equation}
\metric = \transpose{J}(\bundle)J(\bundle),
\end{equation}
and the physically meaningful squared speed is
\begin{equation}
\|\dot{x}\|_2^2
=
\transpose{\bundledot}\transpose{J}(\bundle)J(\bundle)\bundledot
=
\transpose{\bundledot}\metric\bundledot.
\end{equation}
As shown in Fig.~\ref{fig:idea}(b), the same coordinate increment can correspond to different ambient motions depending on configuration. This dependency is why a Euclidean norm on \( \bundledot \) is not geometrically meaningful, whereas \( \|\bundledot\|_{\metric} \) measures the actual ambient motion.

For inertia-dominated systems, the natural metric is the inertia matrix, and the kinetic energy is
\begin{equation}
\kineticenergy(\bundle,\bundledot)
=
\frac{1}{2}\transpose{\bundledot}\metric\bundledot.
\end{equation}
For drag-dominated systems, the natural metric is the drag matrix, and the drag dissipation power is
\begin{equation}
\power(\bundle,\bundledot)
=
\transpose{\bundledot}\metric\bundledot.
\end{equation}
In both cases, \( \metric \) defines the physically meaningful norm of generalized velocity.

Generalized forces are covectors, i.e., elements of \( T_\bundle^*\bundlespace \), that pair with generalized velocities to produce power:
\begin{equation}
\power = \lagrangeforce(\bundledot) = \langle \lagrangeforce ; \bundledot \rangle.
\end{equation}
This vector--covector duality means that a generalized force specifies the power associated with unit generalized velocity in a given direction, as illustrated in Fig.~\ref{fig:idea}(c).

The metric \(\metric\) on \(T_{\bundle}\bundlespace\) canonically induces a dual metric \(\metric^{*}\) on \(T_{\bundle}^{*}\bundlespace\), as illustrated in Fig.~\ref{fig:idea}(c). In a coordinate chart, the positive-definite velocity-metric matrix \(\metric(\bundle)\) induces the dual metric and associated dual norm
\begin{equation} \label{eq:dual_norm}
\metric^*(\bundle) = \inv{\metric}(\bundle),
\qquad
\|\lagrangeforce\|_{\metric^*}
:=
\sqrt{\transpose{\lagrangeforce}\metric^*\lagrangeforce}.
\end{equation}
The dual metric \(\metric^{*}\) is the natural metric on generalized-force space compatible with the velocity metric \( \metric \) through vector--covector duality. It measures force residuals independently of their coordinate representation \cite{ref076,ref081}.

Returning to the pan--tilt example, let
\begin{equation}
\lagrangeforce = \lagrangeforce_\theta d\theta + \lagrangeforce_\phi d\phi
\end{equation}
denote the generalized torques. If torque residuals are measured using the Euclidean coordinate norm, then their relative importance depends on the chosen coordinate chart and scaling. Rescaling a coordinate changes the numerical representation of the torque covector, and the same coordinate norm can correspond to different physical power at different configurations, as suggested by Fig.~\ref{fig:idea}(b) and Fig.~\ref{fig:idea}(d). By contrast, the dual norm in~\eqref{eq:dual_norm} evaluates the covector using the geometry induced by the ambient mechanical space. In this way, the dual metric pulls the coordinate force back into the underlying physical space, just as \( \metric \) pulls the coordinate velocity back into that space.

Define the inverse-dynamics residual at sample \( \sampleidx \) as
\begin{equation}
r_\sampleidx(\dynpar)
:=
\regressor_\sampleidx \dynpar - \lagrangeforce_\sampleidx.
\end{equation}
Standard least squares minimizes the Euclidean coordinate norm \( \|r_\sampleidx(\dynpar)\|_2^2 \), which depends on the coordinate chart. To measure force residuals in a coordinate-independent manner, we instead weight them by the dual metric:
\begin{equation}
\hat{\dynpar}
=
\arg\min_{\dynpar}
\;
\sum_{\sampleidx=1}^{\samplenum}
\transpose{r_\sampleidx(\dynpar)}
\metric^*_\sampleidx(\dynpar)
r_\sampleidx(\dynpar).
\label{eq:dual_metric_ls}
\end{equation}
where \( \metric^*_\sampleidx \) is the system dual metric evaluated at sample \( \sampleidx \). For inertia-dominated systems, \( \metric_\sampleidx \) is the composite inertia matrix; for drag-dominated systems, \( \metric_\sampleidx \) is the composite drag matrix. Equation~\eqref{eq:dual_metric_ls} therefore defines a coordinate-independent identification objective: it measures force residuals through their physical power interpretation rather than through their raw coordinate representation.\footnote{Note that some system identification problems may not be fully observable \cite{ref556}. Nevertheless, because the composite metric tensor typically appears directly in the inverse dynamics, dual-metric normalization preserves the same invariant set of dynamic parameters as the original regression. Therefore, the proposed method remains compatible with systems that are not fully identifiable, such as the Crazyflie--pendulum system studied here.}\footnote{From a statistical perspective, dual-metric weighting can be interpreted as a coordinate-independent default normalization corresponding to an isotropic residual model in metric-normalized force coordinates.}

\subsection{Convex Reformulation via Schur Complements}

At first glance, \eqref{eq:dual_metric_ls} appears difficult to optimize because both the residual \( r_\sampleidx(\dynpar) \) and the weighting matrix \( \metric^*_\sampleidx(\dynpar) \) depend on the unknown parameters \( \dynpar \). Nevertheless, the problem admits a convex reformulation.

For the system classes considered here, the metric is affine in the dynamic parameters:
\begin{equation}
\metric_\sampleidx(\dynpar)
=
\metric_{\sampleidx,0}
+
\sum_{\paramidx=1}^{\dynpardim}
\dynpar_{\paramidx}\metric_{\sampleidx,\paramidx},
\label{eq:metric_affine}
\end{equation}
where each \( \metric_{\sampleidx,\paramidx}\in\mathbb{S}^{\gendim} \) is known for sample \( \sampleidx \), and \( \metric_\sampleidx\in\mathbb{S}_{++}^{\gendim} \) is enforced through physical consistency constraints. This affine dependence arises because all nonlinearities enter through the configuration variables and their associated Jacobians, rather than through the dynamic parameters. The condition includes, but is not limited to, the inertia and drag coefficients discussed in \eqref{eq:inertia} and \eqref{eq:drag}; it also applies to other Riemannian metrics. Likewise,
\begin{equation}
r_\sampleidx(\dynpar)=\regressor_\sampleidx\dynpar-\lagrangeforce_\sampleidx
\end{equation}
is affine in \( \dynpar \). Thus, the only nonlinearity in \eqref{eq:dual_metric_ls} is the quadratic form involving the inverse of an affine matrix.

For each sample, introduce a slack variable \( s_\sampleidx \in \euclid \) and impose the epigraph bound
\begin{equation}
s_\sampleidx
\ge
\transpose{r_\sampleidx(\dynpar)}
\metric^*_\sampleidx(\dynpar)
r_\sampleidx(\dynpar).
\label{eq:epigraph_bound}
\end{equation}
By the Schur complement, \eqref{eq:epigraph_bound} is equivalent to
\begin{equation}
\begin{bmatrix}
\metric_\sampleidx(\dynpar) & r_\sampleidx(\dynpar) \\
\transpose{r_\sampleidx(\dynpar)} & s_\sampleidx
\end{bmatrix}
\succeq 0,
\label{eq:schur_block}
\end{equation}
provided that \( \metric_\sampleidx\succ 0 \). Because both blocks are affine in \( \dynpar \), \eqref{eq:schur_block} is a convex semidefinite constraint.

\begin{figure}[!t]
\centering
\includegraphics[width=\linewidth]{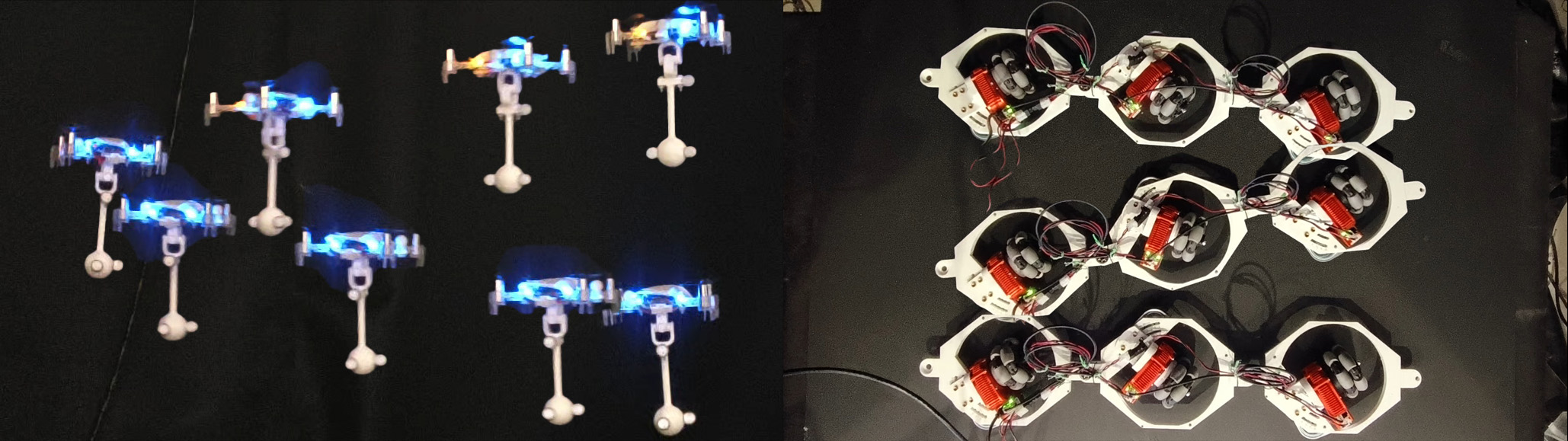}
\caption{Experimental platforms: the inertia-dominated Crazyflie with a pendulum attached through a universal joint (left), and the drag-dominated three-link \censor{Land}Salp robot (right), together with snapshots of representative execution trajectories.}
\label{fig:experiment}
\end{figure}

The original problem is therefore equivalent to
\begin{equation}
\hat{\dynpar}
=
\arg\min_{\dynpar,\{s_\sampleidx\}}
\sum_{\sampleidx=1}^{\samplenum} s_\sampleidx
\quad
\text{s.t.}\;
\eqref{eq:schur_block}\ \forall \sampleidx.
\label{eq:convex_dual_metric}
\end{equation}
The objective is linear in the slack variables, and all constraints are convex. Hence, \eqref{eq:convex_dual_metric} remains a convex semidefinite program.

This reformulation shows that coordinate-independent system identification does not require giving up the computational advantages of inverse-dynamics least squares. Although the dual metric depends on the unknown parameters, its affine structure allows the weighted residual to be represented exactly through Schur-complement LMIs. The proposed method therefore preserves the practical tractability of convex identification while replacing the coordinate-dependent Euclidean force residual with a geometrically correct, coordinate-independent one.

We solve \eqref{eq:convex_dual_metric} in dual form: for \(N\) samples, \(n\) generalized velocities, and \(p\) model parameters, this formulation has \(N\) positive-semidefinite matrix variables of order \(n+1\), together with \(N+p\) linear equality constraints. This structure keeps the proposed method practical at the dataset scales considered here. We report solve times using MOSEK on an Intel Core i5-14400F CPU; physical-consistency constraints are included for both the proposed method and WLS. For Crazyflie (\(N=21{,}204\), \(n=8\)), the proposed method requires \(12.4\,\mathrm{s}\), compared with \(6.5\,\mathrm{s}\) for WLS. For LandSalp (\(N=38{,}413\), \(n=5\)), it requires \(19.5\,\mathrm{s}\), compared with \(3.9\,\mathrm{s}\) for WLS. Thus, the proposed formulation incurs a moderate offline solve-time overhead in exchange for coordinate invariance. For general complexity results on semidefinite programming, we refer the reader to \cite{ref605}.

\section{Experiments}

\begin{figure}[!t]
\centering
\includegraphics[width=\linewidth]{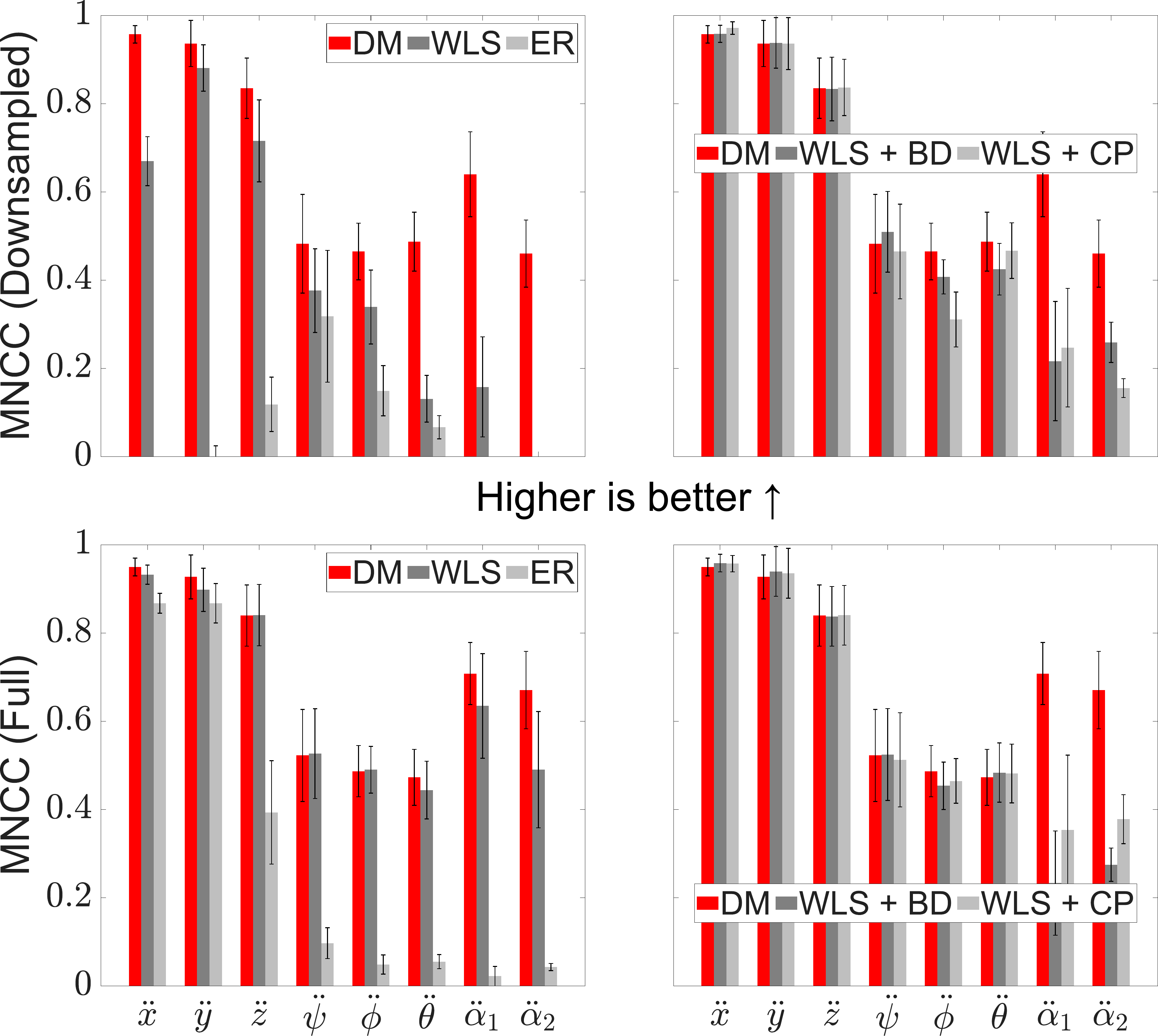}
\caption{MNCC between predicted and measured forward dynamics for the Crazyflie--pendulum system on the test set. Top/bottom: models trained on downsampled/full data. Left: comparison with WLS and energy-based least squares. Right: comparison with regularized WLS using Bregman divergence and constant pullback. Higher is better; error bars denote test-set standard deviation. The proposed method (DM) consistently outperforms the alternatives, especially on the shape coordinates (\( \ddot{\alpha}_1 \) and \( \ddot{\alpha}_2 \))}
\label{fig:crazyflie}
\end{figure}

We evaluated the proposed method on two robotic systems with different dynamics: a Crazyflie 2.0 quadrotor with a pendulum attached through a universal joint, and the \censor{Land}Salp robot (Fig.~\ref{fig:experiment}). These systems are, respectively, inertia-dominated and viscous drag-dominated, matching the two classes of dynamics studied in this paper. They also provide challenging testbeds for coordinate-dependent identification: their generalized coordinates have different units and scales, admit multiple coordinate choices, and induce configuration-dependent inner products under the corresponding system metrics. For floating-base coordinates, we use body velocity and body-velocity change rate as \( \bundledot \) and \( \bundleddot \), respectively.

For both platforms, we recorded data using an OptiTrack motion capture system together with onboard sensing. We compared the proposed dual-metric-WLS against error-covariance-WLS, energy-based least squares, and two regularized WLS variants using Bregman-divergence and constant-pullback regularizers. For the regularized methods, the error covariance is first estimated from the corresponding unregularized problem, following \cite{ref325}. All methods were required to satisfy the same physical-consistency constraints.

We evaluated all methods in a practical downsampled dataset,\footnote{Downsampling is applied after preprocessing and regressor construction, and the downsampled identification regression remains the same rank as the full-data regression for both systems.} and the complete training dataset. All evaluations use a held-out test dataset distinct from the training data. Given the measured states and generalized forces, forward dynamics predicts accelerations for Crazyflie--pendulum and velocities for LandSalp. We report the maximum normalized cross-correlation (MNCC) between the predicted signals $\hat{s}$ and measured signals $s$:
\begin{equation}
\rho_{\max}(s, \hat{s})
=
\max_{|\ell|\leq \ell_{\max}}
\frac{\sum_t \hat{s}_{t+\ell}s_t}
{\sqrt{\sum_t \hat{s}_t^2\sum_t s_t^2}},
\end{equation}
where \(\ell_{\max}\) corresponds to a phase shift of \(5^\circ\) of the slowest motion period. Because the forward-dynamics measurements contain high-frequency oscillations, MNCC provides a more appropriate measure of similarity in temporal alignment and waveform shape than pointwise error metrics such as RMSE. For a more direct comparison, we also plot representative predicted and measured trajectories for the coordinates exhibiting the largest performance differences.

\begin{figure}[!t]
\centering
\includegraphics[width=\linewidth]{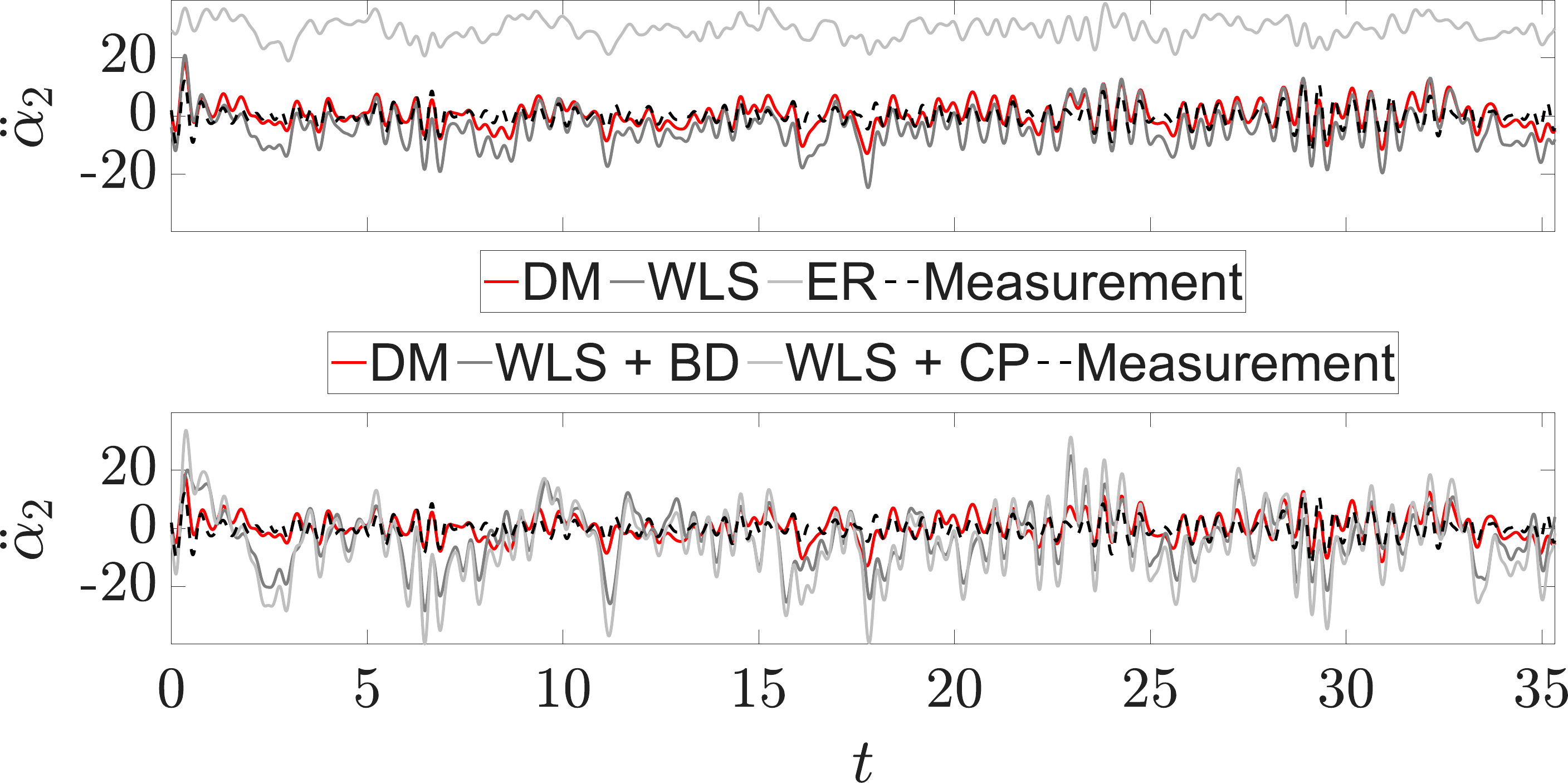}
\caption{Predicted and measured \(\ddot{\alpha}_2\) for the Crazyflie--pendulum system, comparing models trained on the full dataset using the proposed method, WLS, energy-based least squares, and the two regularized WLS variants.}
\label{fig:crazyflie_traj}
\end{figure}

\subsection{Inertia-Dominated System}

The inertia-dominated platform is a Crazyflie 2.0 with a pendulum attached through a universal joint. Based on prior work \cite{ref503}, the system is differentially flat, with flat outputs given by the quadrotor \(x\)-, \(y\)-, and \(z\)-positions and yaw. We generated excitation trajectories using sinusoidal motions with mutually non-harmonic frequencies so that the full trajectory did not repeat over the experiment duration.

In addition to motion-capture measurements, we recorded onboard IMU data, battery voltage, and PWM commands to the four thrusters, all at \(100\) Hz. Generalized forces were computed from PWM and battery voltage using firmware calibration values. We collected six trajectories for training and six for testing; each lasted about \(35\) s, and each test trajectory started from a different phase. The identified parameters include the mass, center-of-mass moments, and inertia tensors of the quadrotor body, universal-joint core, and pendulum. Nominal values for the regularized baselines were obtained from the CAD model and drone datasheet.

For the downsampled dataset, we used about \(20\) samples from the processed training set. As shown in the top panel of Fig.~\ref{fig:crazyflie}, the proposed method significantly outperformed WLS and the energy-based method, with the largest gains in body angular accelerations and pendulum shape accelerations. Compared with the regularized baselines, it achieved similar performance on body linear and angular coordinates while still outperforming them on the pendulum shape coordinates.

Using the full training set, most methods performed similarly on body translation and attitude, but the proposed method still outperformed all alternatives on the two pendulum shape coordinates (bottom panel of Fig.~\ref{fig:crazyflie}). Thus, even when covariance weighting is well supported statistically, coordinate-independent weighting remains beneficial for internal shape dynamics.

Figure~\ref{fig:crazyflie_traj} shows a representative pendulum pitch-shape-acceleration trajectory. The proposed method captures most of the motion, whereas WLS and the energy-based method produce incorrect predictions. This result reflects the central geometric issue: residuals measured directly in the gimbal coordinates are distorted by the nonlinear coordinate representation, while the dual metric pulls them back into the ambient mechanical space.

\begin{figure}[!t]
\centering
\includegraphics[width=\linewidth]{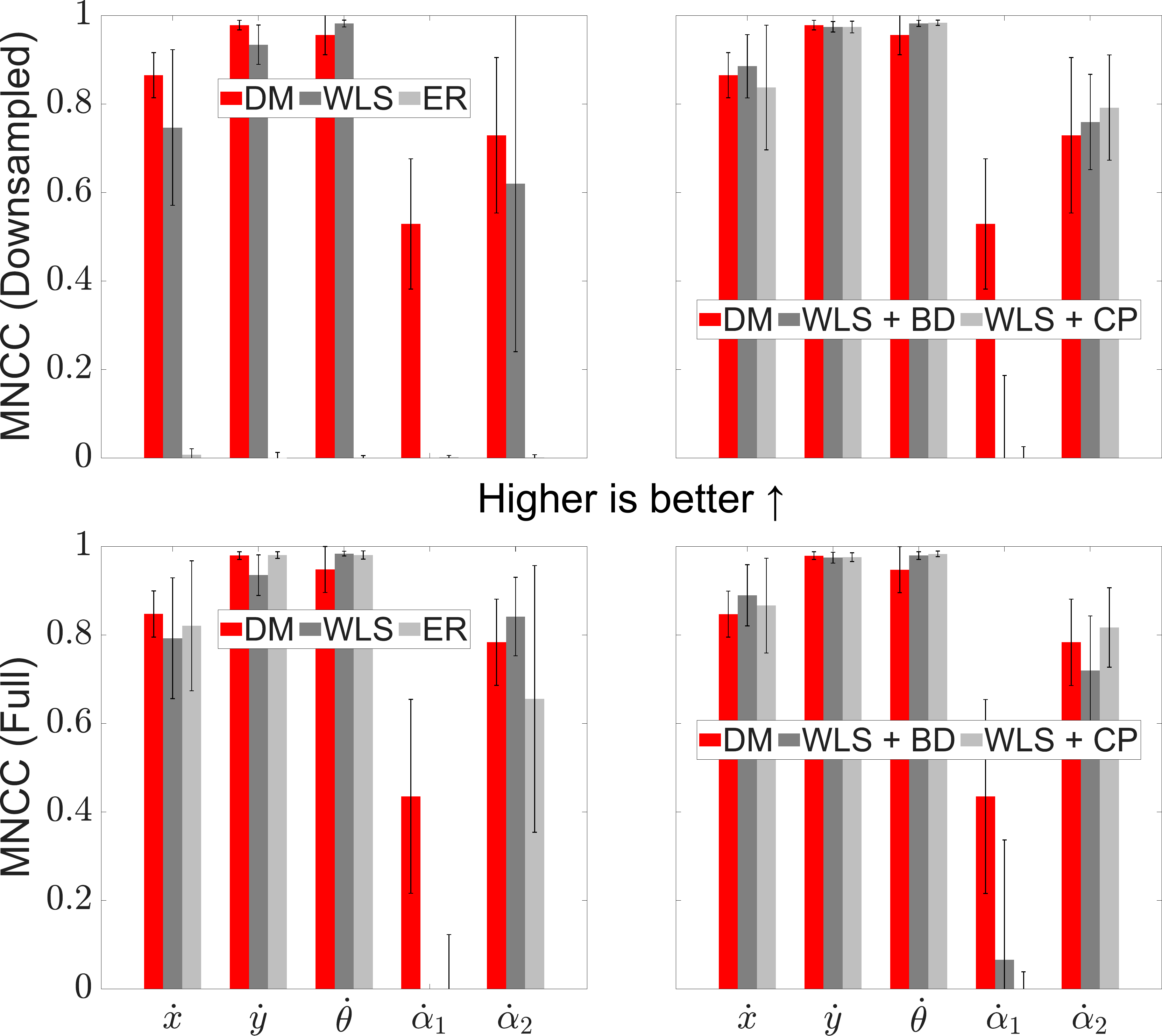}
\caption{MNCC between predicted and measured forward dynamics for the \censor{Land}Salp system on the test set. Top/bottom: models trained on downsampled/full data. Left: comparison with WLS and energy-based least squares. Right: comparison with regularized WLS using Bregman divergence and constant pullback. Higher is better; error bars denote test-set standard deviation. The proposed method (DM) consistently outperforms the alternatives, especially on the shape coordinates (\( \dot{\alpha}_1 \) and \( \dot{\alpha}_2 \))}
\label{fig:landsalp}
\end{figure}

\subsection{Drag-Dominated System}

The drag-dominated platform is the \censor{Land}Salp robot, a multibody locomotion system with three omni wheels driven by series elastic actuators and two passive joints. The training trajectory consisted of eight excitation segments totaling about three minutes, generated by driving the three wheels through distinct sinusoidal combinations. Test trajectories consisted of forward, backward, leftward, rightward, and turning gaits from prior work \cite{ref586}. In addition to motion-capture data, we recorded actuator force measurements.

The identified parameters include viscous drag coefficients for each body unit and joint, as well as the internal drag coefficient of each actuator. Identification was performed in two stages: actuator identification from output force, actual velocity, and commanded velocity; then link and joint drag identification using the inverse-dynamics framework in this paper. Nominal values for the regularized baselines were taken from prior identification results \cite{ref586}.

For the downsampled dataset, we used about \(40\) samples from the processed training set. As shown in the top panel of Fig.~\ref{fig:landsalp}, the proposed method outperformed WLS, the energy-based method, and the regularized baselines on the shape coordinates. The largest contrast appeared in the first shape coordinate, where all methods except DM exhibited non-positive cross-correlation, indicating predictions roughly \(90^\circ\) to \(180^\circ\) out of phase with the measurements.

With the full training set (bottom panel of Fig.~\ref{fig:landsalp}), the baselines performed similarly on body position and orientation, but the proposed method still outperformed them on the shape coordinates.

\begin{figure}[!t]
\centering
\includegraphics[width=\linewidth]{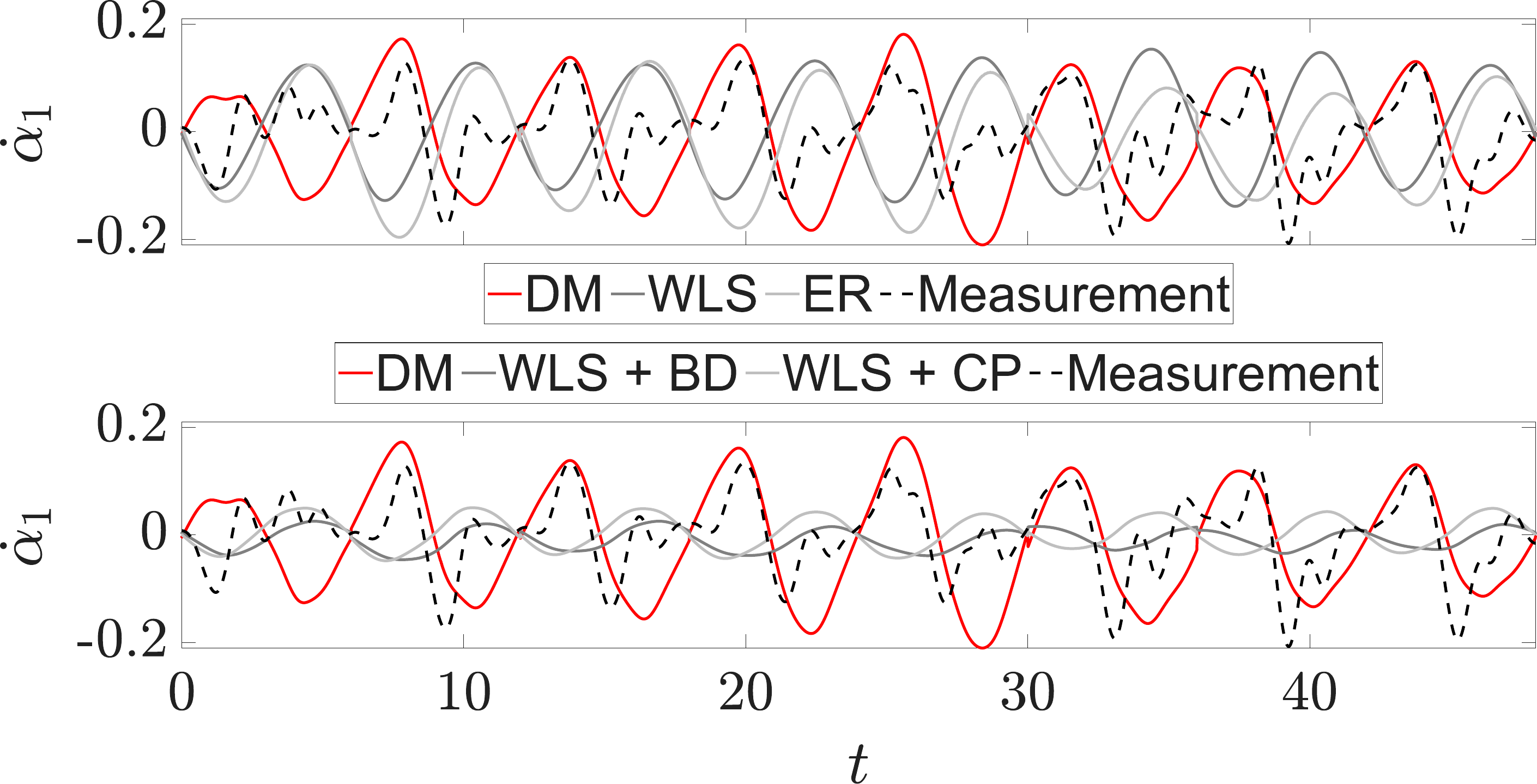}
\caption{Predicted and measured \(\dot{\alpha}_1\) for the \censor{Land}Salp system, comparing models trained on the full dataset using the proposed method, WLS, energy-based least squares, and the two regularized WLS variants.}
\label{fig:landsalp_traj}
\end{figure}

Figure~\ref{fig:landsalp_traj} shows a representative first-joint-velocity trajectory during leftward motion. The proposed method captures most of the joint motion, whereas WLS and the energy-based method show large phase errors. As in the Crazyflie experiment, this result indicates that residuals measured directly in joint coordinates are distorted by the nonlinear coordinate representation, while dual-metric weighting recovers the underlying physical information more faithfully.

\subsection{Discussion}

Despite noisy sensor measurements, approximate actuation forces, and model simplifications, the proposed method consistently performed best on both downsampled and full dataset. This robustness suggests that its advantage is geometric rather than merely statistical: by pulling residuals back into the ambient mechanical space, the dual metric removes coordinate dependence and improves data efficiency. The improved data efficiency may arise because coordinate-independent normalization maps samples into a common ambient-space geometry, thereby better revealing each sample's contribution to parameter identifiability.

By contrast, the energy-based method consistently performed poorly, likely because its effectively rank-one weighting compresses vector residual information into a single scalar channel, reducing identifiability and worsening conditioning. WLS improved only when enough data were available to estimate the residual covariance reliably, yet it remained less effective on shape coordinates, where nonlinear coordinate representations are most severe. The regularized baselines benefited from additional prior information, but they still inherited the coordinate dependence of the underlying residual and may be less reliable in practice when nominal parameters are unavailable or inaccurate.

In addition to improved accuracy, the proposed method was also numerically more robust in practice. The dual metric acts as a physically meaningful self-scaling mechanism, substantially reducing the need for manual scaling and numerical tuning compared with the competing methods.

\section{Conclusion and Future Work}

This paper proposed a coordinate-independent system identification method by applying dual metrics as weights in inverse-dynamics least squares. By using the dual metric induced by the system Riemannian metric, the method measures residuals in a physically meaningful way, pulling coordinate force residuals back into the ambient mechanical space and eliminating dependence on coordinate choice, units, and scaling. The resulting objective remains a convex optimization problem through an affine-metric and Schur-complement reformulation, while remaining compatible with physical-consistency constraints and geometric regularization. Experiments on an inertia-dominated Crazyflie--pendulum system and a drag-dominated \censor{Land}Salp robot showed the best average performance across both downsampled and full datasets, with the largest gains on shape coordinates, where nonlinear coordinate representations make coordinate dependence most severe. These results show that coordinate independence is not only a geometric property, but also a practical source of robustness and data efficiency in robot model identification.

Future work includes extending this coordinate-independent viewpoint to broader identification settings in which time-domain data on manifolds is used, such as neural ODE or neural SDE models \cite{ref096}. Another direction is online parameter estimation, where the dual-metric weighting may need efficient updates or approximations during adaptation \cite{ref606}. It is also important to study systems governed by multiple metrics, especially when different physical effects dominate on different time scales. Developing trajectory- or controller-aware excitation strategies that work in conjunction with dual-metric weighting to improve identifiability is another promising direction. Finally, incorporating sample independence into the objective may improve robustness to repeated or redundant data and clarify its relationship to identifiability.

\bibliographystyle{IEEEtran}
\bibliography{ref}

\end{document}